\documentclass[10pt,twocolumn,letterpaper]{article}

\usepackage{iccv}
\usepackage{times}
\usepackage{epsfig}
\usepackage{graphicx}
\usepackage{amsmath}
\usepackage{amssymb}

\usepackage{multirow} 
\usepackage{enumerate}
\usepackage{booktabs}
\usepackage{comment}
\usepackage{subfigure}
\usepackage[breaklinks=true,bookmarks=false]{hyperref}

\iccvfinalcopy 

\def\iccvPaperID{4506} 

\newcommand {\mymodel}{Non-local Recurrent Neural Memory\xspace}
\begin{document}

\title{\mymodel for Supervised Sequence Modeling}

\author{Canmiao Fu\textsuperscript{1,2,}\footnotemark[1],\ \ \ Wenjie Pei\textsuperscript{2,}\footnotemark[1],\ \ \ Qiong Cao\textsuperscript{2},\ \ \ Chaopeng Zhang\textsuperscript{1}, 
\\Yong Zhao\textsuperscript{1,}\footnotemark[2], \ \ Xiaoyong Shen\textsuperscript{2}\ \ and 
Yu-Wing Tai\textsuperscript{2,}\footnotemark[2]\vspace{1mm} \\
\normalsize\textsuperscript{1}School of ECE, Peking University\quad	\textsuperscript{2}Tencent\\
	{\tt \footnotesize fcm@pku.edu.cn, wenjiecoder@outlook.com, freyaqcao@tencent.com, cpz@pku.edu.cn,\vspace{-1mm}}\\
		{\tt\vspace{-2mm}\footnotesize yongzhao@pku.edu.cn, goodshenxy@gmail.com, yuwingtai@tencent.com\vspace{-0.05in}}
}

\maketitle
\renewcommand{\thefootnote}{\fnsymbol{footnote}}
\footnotetext[1]{Both authors contributed equally.} 
\footnotetext[2]{Corresponding authors.}

\renewcommand{\thefootnote}{\arabic{footnote}}


\begin{abstract}


\vspace{-0.1in}
Typical methods for supervised sequence modeling are built upon
the recurrent neural networks to capture temporal dependencies. 
One potential limitation of these methods is that they only model explicitly 
information interactions between adjacent time steps in a sequence, 
hence the high-order interactions between nonadjacent time steps are not fully 
exploited. It greatly limits the capability of modeling the long-range temporal dependencies since one-order interactions cannot be maintained for a long term due to information dilution and gradient vanishing. To tackle this limitation, we propose the Non-local Recurrent Neural Memory (NRNM) for supervised sequence modeling, which performs non-local operations to learn full-order interactions within a sliding temporal block and models global interactions between blocks in a gated recurrent manner. Consequently, our model is able to capture the long-range dependencies. Besides, the latent high-level features contained in high-order interactions can be distilled by our model. We demonstrate the merits of our NRNM on two different tasks: action recognition and sentiment analysis.
\vspace{-0.15in}
\end{abstract}

\section{Introduction}
\vspace{-0.07in}
Supervised sequence modeling aims to build models to extract effective features from variety of sequence data such as 
video data or text data via supervised learning. It has extensive applications ranging from computer 
vision~\cite{pei2017temporal, shahroudy2016ntu} to 
natural language processing~\cite{grave2016improving,vaswani2017attention}. 
The key challenge in supervised sequence modeling is to capture the long-range temporal dependencies, which are used to further 
learn the high-level feature for the whole sequence.

\begin{figure}[t]
	\begin{center}
	\includegraphics[width=0.95\linewidth]{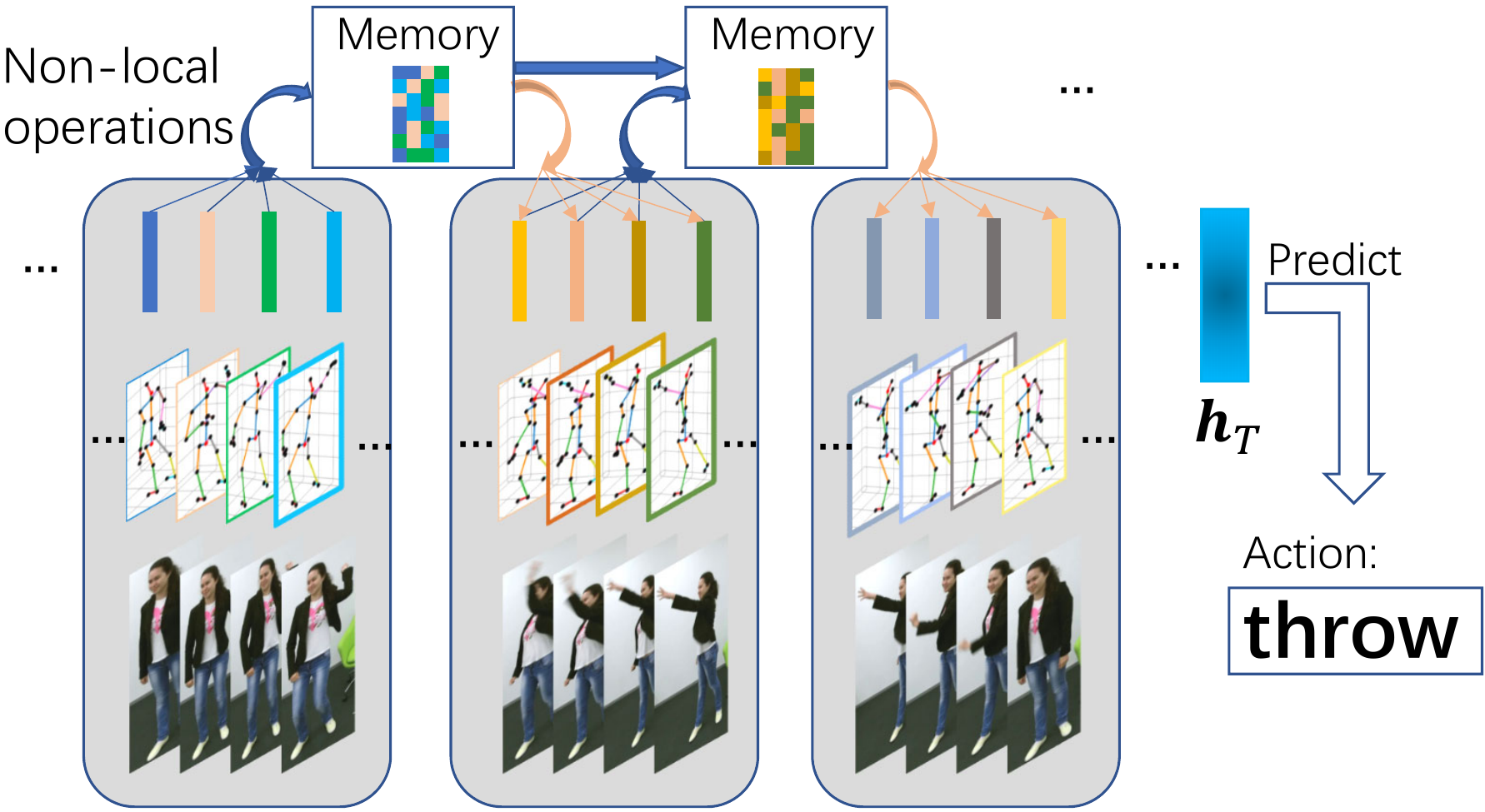}
	\end{center}
	\caption{Given a video sample for action recognition, our proposed model (\emph{NRNM}) performs non-local operations within each memory block to learn high-order interactions between hidden states of different time steps. Meanwhile, the global interactions between memory blocks are modeled in a gated recurrent manner. The learned memory states are in turn leveraged to refine the hidden states in future time steps. Thus, the long-range dependencies can be captured. The model predicts the action based on the hidden state ($\mathbf{h}_T$) in the last time step.}
	\vspace{-0.2in}
	\label{fig:intro}
	\end{figure}
	
Most state-of-the-art methods for supervised sequence modeling are built upon the recurrent neural networks (RNN)~\cite{rumelhart1988learning}, 
which has been validated its effectiveness~\cite{sak2014long,zhang2018adding}. One crucial limitation of the vanila-RNNs is the gradient-vanishing problem 
along the temporal domain, which results in the inability to model long-term dependencies. This limitation is then substantially mitigated 
by gated recurrent networks such as GRU~\cite{cho2014learning} and LSTM~\cite{hochreiter1997long}, which employ learnable gates to selectively 
retain information in the memory or hidden states. 
The memory-based methods for sequence modeling~\cite{santoro2018relational, sukhbaatar2015end,weston2014memory} are further proposed 
to address the issue of limited memory of recurrent networks. However, a potential drawback of these methods is that they only model 
explicitly the information interactions between adjacent time steps in the sequence, hence the high-order interactions 
between nonadjacent time steps are not fully exploited. This drawback gives rise to two negative consequences: 
1) the high-level features contained in the interactions between nonadjacent time steps cannot be distilled; 
2) it greatly limits the modeling of long-range temporal dependencies since one-order interaction information cannot be maintained 
in a long term due to the information dilution and gradient vanishing along with recurrent operations.

Inspired by non-local methods~\cite{buades2005non, wang2018non} which aim to explore potential interactions between all pairs of 
feature portions, we propose to perform non-local operations to model the high-order interactions between non-adjacent time steps in a 
sequence. The captured high-order interactions are able to not only help to distill latent high-level 
features which are hardly learned 
by typical sequence modeling methods focusing on one-order interactions, but also contribute to modeling long-range temporal dependencies since the non-local operations strength the latent feature propagation and thus substantially alleviate the vanishing-gradient problem.
Since exploring full-order interactions between all time steps for a long sequence is computationally expensive and also not necessary 
due to information redundancy, we model full-order interactions by non-local operations within a temporal block (a segment of sequence) and 
slide the block to recurrently update the extracted information.
 More specifically, we propose the Non-local Recurrent Neural Memory (\emph{NRNM}) to perform the blockwise non-local operations to 
 learn full-order interactions within each memory block and capture the local but high-resolution temporal dependencies. 
 Meanwhile, the global interactions between adjacent blocks are captured by updating the memory states 
 in a gated recurrent manner when sliding the memory cell. Consequently, the long-range dependencies can be maintained. Figure~\ref{fig:intro} illustrates our method by an example of action recognition.

Compared to typical supervised sequence modeling methods, especially recurrent networks with memory mechanism, 
our \emph{NRNM} benefits from following advantages:
\begin{itemize}
	\vspace{-0.1in}
\item It is able to model 1) the local full-order interactions between all time steps within a segment of sequence (memory block) and 
2) the global interactions between memory blocks. Thus, It can capture much longer temporal dependencies. 
\vspace{-0.1in}
\item The proposed \emph{NRNM} is able to learn latent high-level features contained in high-order interactions between non-adjacent time steps, 
which may be missed by conventional methods.
\vspace{-0.1in}
\item The \emph{NRNM} cell can be seamlessly integrated into any existing sequence models with recurrent structure to enhance the power of
 sequence modeling. The integrated model can be trained in an end-to-end manner.
\end{itemize}

\vspace{-0.2in}
\section{Related work}
\vspace{-0.08in}

\noindent\textbf{Graphical sequence models.}
The conventional graphical models for sequence modeling can be roughly divided into two categories: generative and discriminative models. A well-known example of generative model is Hidden Markov Model (HMM)~\cite{rabiner1989tutorial}, which models sequence data in a chain of latent $k$-nomial features. 
Discriminative graphical models model the distribution over all class labels conditioned on the input data.
Conditional Random Fields (CRF)~\cite{lafferty2001conditional} is a discriminative model for sequential predictions by modeling the linear mapping between observations and labels. To tackle its limitation of linear mapping, many nonlinear CRF-variants are proposed~\cite{morency2007latent, pei2018multivariate, peng2009conditional, van2011hidden}.
The disadvantages of graphical model compared to recurrent networks lie in the hard optimization and limited capability of temporal modeling. Our model is designed based on recurrent networks. 


\noindent\textbf{Recurrent Networks.} 
Recurrent Neural Network~\cite{rumelhart1988learning} learns a hidden representation for each time step by taking into account 
both current and previous information. Benefited from its advantages such as easy training and temporal modeling, it has been successfully applied to, amongst others,  
handwriting recognition~\cite{bertolami2009novel} and speech recognition~\cite{sak2014long}. However, the key limitation of vanila-RNN is the gradient vanishing problem during training~\cite{hochreiter2001gradient} and thus cannot model long-range temporal dependencies.
This limitation is alleviated by gated recurrent networks such as Long Shot-Term Memory (LSTM)~\cite{hochreiter1997long} and Gate Recurrent Unit (GRU)~\cite{cho2014learning}, which selectively retain information by learnable gates. Nevertheless, a potential limitation of these models is that they only model explicitly one-order interactions between adjacent time steps, hence the high-order interactions between nonadjacent time steps are not fully captured.
Our model is proposed to circumvent this drawback by employing non-local operations to model full-order interactions in a block-wise manner. Meanwhile, the global interactions between blocks are modeled by a gated recurrent mechanism. Thus, our model is able to model long-range temporal dependencies and distill high-level features that are contained in high-order interactions.

\noindent\textbf{Memory-based recurrent networks.}
Memory networks are first proposed to rectify the drawback of limited memory of recurrent networks~\cite{sukhbaatar2015end, weston2014memory}, 
which are then extended for various tasks, especially in natural language processing. Most of these models build external memory units upon a basis model to augment its 
memory~\cite{graves2014neural, santoro2018relational, sukhbaatar2015end, weston2014memory}.  
In particular, attention mechanism~\cite{bahdanau2014neural} is employed to filter the information flow from memory~\cite{grave2016improving, kumar2016ask, sukhbaatar2015end, xiong2016dynamic}. 
The primary difference between these memory-based recurrent networks and our model is that these models focus on augmenting the memory size to memorize more information for reference while our model aims to model high-order interactions between different time steps in a sequence, which is not concerned by existing memory-based networks.

\vspace{-0.1in}
\section{\mymodel}
\vspace{-0.05in}
\begin{figure*}[tb]
	\begin{center}
		\includegraphics[width=0.85\linewidth]{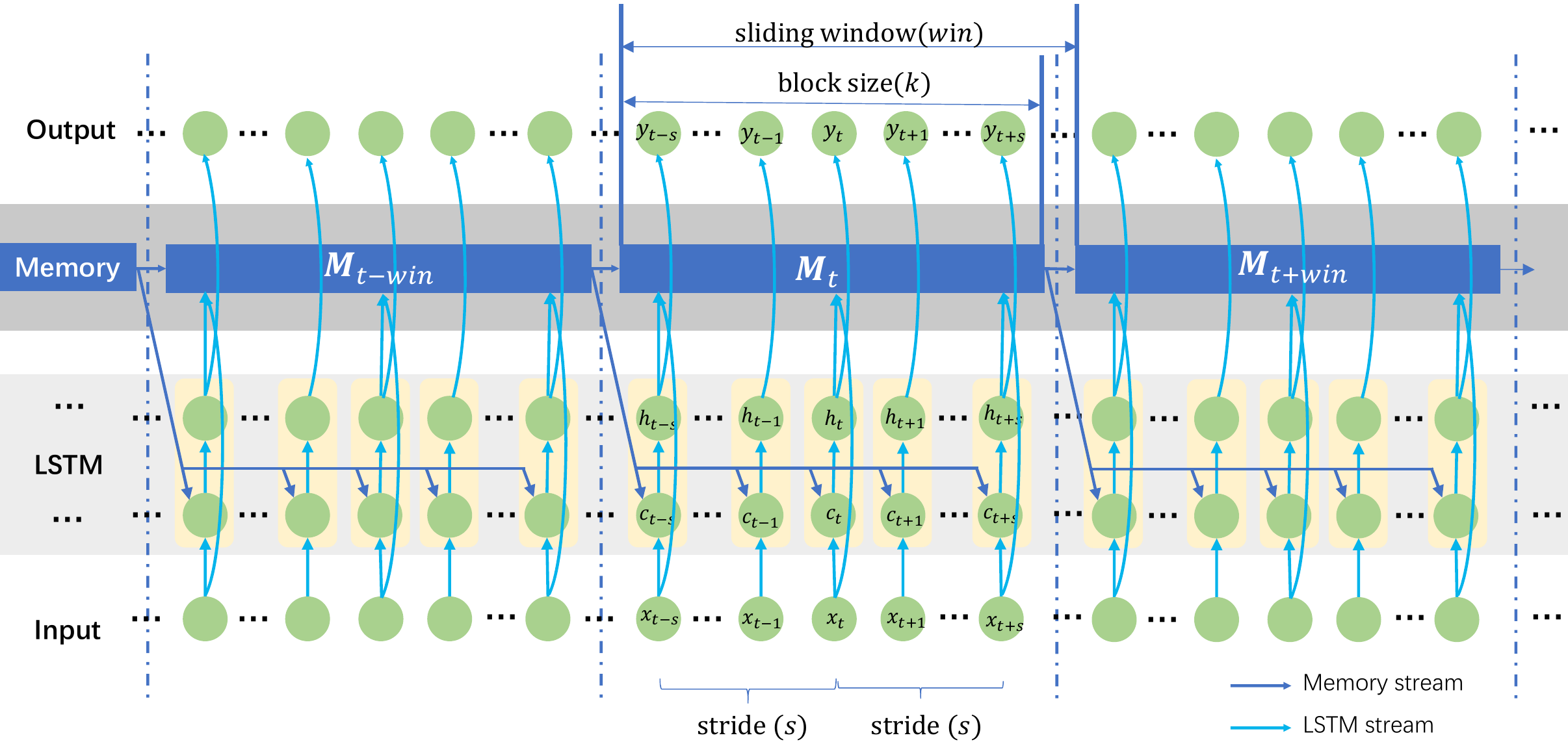}
	\end{center}
	\vspace{-0.1in}
	\caption{The architecture of our method. Our proposed \emph{NRNM} is built upon the LSTM backbone to learn high-order interactions between LSTM hidden states of different time steps within each memory block. Meanwhile, the global interactions between memory blocks are modeled in a gated recurrent manner. The learned memory states are in turn used to refine the LSTM hidden states in future time steps. }
	\label{fig:model}
	\vspace{-0.15in}
\end{figure*}

Given as input a sequence, our Non-local Recurrent Neural Memory (\emph{NRNM}) is designed as a 
memory module to capture the long-range temporal dependencies in a non-local manner. 
It can be seamlessly integrated into any existing sequential models with recurrent structure to enhance 
sequence modeling. As illustrated in Figure~\ref{fig:model}, we build our \emph{NRNM} upon a LSTM backbone as an 
instantiation. We will first elaborate on the cell structure of our \emph{NRNM} and then describe how \emph{NRNM} and the LSTM backbone perform sequence modeling collaboratively. 
\vspace{-0.05in}
\subsection{NRNM Cell}
\vspace{-0.05in}
\label{sec:nrnm}
Residing on a basis sequence model (a standard LSTM backbone), the proposed Non-local Recurrent Neural Memory (\emph{NRNM}) aims to maintain 
a memory cell along the temporal dimension which can not only distill the underlying information contained in past time steps of the input 
sequence but also capture the temporal dependencies, both operating in the long temporal range. 
To this end, the \emph{NRNM} cell performs non-local operations on a segment of the sequence (termed as non-local memory block) 
sliding along the temporal domain, as shown in Figure~\ref{fig:model}. This blocking design is analogous to DenseNet~\cite{huang2017densely} 
which performs dense connection (a form of non-local operation) in blocks. The obtained memory embeddings are further leveraged to refine the hidden
embeddings which are used for the final prediction. The memory state is updated recurrently when sliding the memory block, which is consistent 
with update of the basis LSTM backbone.

Consider an input sequence $\mathbf{x}_{1, \dots, T} = \{\mathbf{x}_1, \dots, \mathbf{x}_T\}$ of length $T$ in 
which $\mathbf{x}_t \in \mathbb{R}^D$ denotes the observation at the $t$-th time step. 
The hidden representation of the input sequence at time step $t$ learned by the LSTM backbone is denoted as $\mathbf{h}_t$. \emph{NRNM} learns the memory embedding $\widetilde{\mathbf{M}}_t$ for a block (a segment of time steps) with blocking size $k$ covering the temporal interval $[t-k+1, t]$ by refining the underlying information contained in this time interval. Specifically, we consider two types of source information for \emph{NRNM} cell: 1) the learned hidden representations in this time interval $[\mathbf{h}_{t-k+1}, \dots, \mathbf{h}_t]$ by the LSTM backbone; 2) the original input features $[\mathbf{x}_{t-k+1}, \dots, \mathbf{x}_t]$. Hence the memory embedding $\widetilde{\mathbf{M}}_t$ at time step $t$ is formulated as:
\vspace{-0.05in}
\begin{equation}
\vspace{-0.05in}
\widetilde{\mathbf{M}}_t = f([\mathbf{h}_{t-k+1}, \dots, \mathbf{h}_t], [\mathbf{x}_{t-k+1}, \dots, \mathbf{x}_t]),
\end{equation}
where $f$ is the nonlinear transformation function performed by \emph{NRNM} cell. Here we incorporate the input feature $\mathbf{x}$ which is already assimilated in the hidden representation $\mathbf{h}$ of the basis LSTM backbone since we aim to explore the latent interactions between hidden representations and input features in the current block (i.e., the interval $[t-k+1, t]$).

\begin{figure}[htb]
	\begin{center}
		\includegraphics[width=0.95\linewidth]{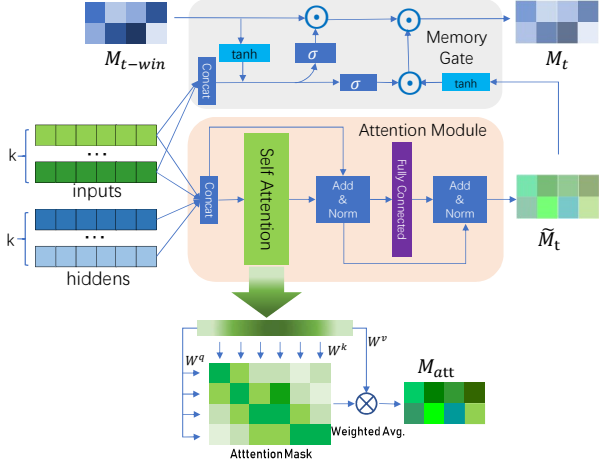}
	\end{center}
	\vspace{-0.1in}
	\caption{The structure of \emph{NRNM} cell.}
	\label{fig:memory}
	\vspace{-0.15in}
\end{figure}

Next we elaborate on the transformation function $f$ of \emph{NRNM} cell presented in Figure~\ref{fig:memory}. 
To distill information in current block that is worth to retain in memory,  we apply Self-Attention 
mechanism implemented with multi-head attention~\cite{vaswani2017attention} to model latent full-order interactions among source informations: original input features and the hidden representations by LSTM in the current block:
\vspace{-0.05in}
\begin{equation}
\vspace{-0.05in}
\begin{split}
&\mathbf{C} = \text{Concat}([\mathbf{h}_{t-k+1}, \dots, \mathbf{h}_t], [\mathbf{x}_{t-k+1}, \dots, \mathbf{x}_t]),\\
& \mathbf{Q, K, V} = (\mathbf{W}^q, \mathbf{W}^k, \mathbf{W}^v) \mathbf{C}, \\
& \mathbf{W}_{att} = \text{softmax}(\mathbf{Q}\mathbf{K^\top}/\sqrt{m}), \\
& \mathbf{M}_{att} = \mathbf{W}_{att} \mathbf{V}.
\label{eqn:transform}
\end{split}
\end{equation}
Herein, $\mathbf{Q, K, V}$ are queries, keys and values of Self-Attention transformed by parameters $\mathbf{W}^q, \mathbf{W}^k, \mathbf{W}^v$ from the 
source information $\mathbf{C}$ respectively. $\mathbf{W}_{att}$ is the derived attention weights calculated by dot-product attention scheme scaled by the 
memory hidden size $m$. The obtained attention embeddings $\mathbf{M}_{att}$ is then fed into two skip-connection layers and one fully-connected layer to achieve the memory embedding $\widetilde{\mathbf{M}}_t$.


The physical interpretation of this design is that the source information is composed of $2k$ information units: $k$ LSTM hidden states and $k$ input features. Each information unit of the obtained memory embedding $\widetilde{\mathbf{M}}_t$ is constructed by attending into each of these $2k$ source information units while the size of memory embedding $\widetilde{\mathbf{M}}_t$ can be customized via the parametric transformation. As such, the full-order latent interactions between the source information units are explored in a non-local way.
Another benefit of such non-local operations is that it strengths latent feature propagation and thus alleviates the vanishing-gradient problem, which is always suffered by recurrent networks.

 Since LSTM hidden states already contain history information by recurrent structure, 
 in practice we use a striding scheme to select hidden states as the source information for \emph{NRNM} cell to avoid potential information redundancy and improve the modeling efficiency. For instance, we pick hidden states every $s$ time steps in the temporal interval $[t-k+1, t]$ for the source information, given $\text{stride} = s$.

\vspace{-0.05in}
\smallskip\noindent\textbf{Gated recurrent update of memory state.} $\widetilde{\mathbf{M}}_t$ only contains information within current temporal block ($[t-k+1, t]$). To model the temporal consistency between adjacent memory blocks, we also update the memory state in a gated recurrent manner, which is similar to the recurrent scheme of LSTM. Specifically, the final memory state $\mathbf{M}_t$ for current memory block is obtained by:
\vspace{-0.07in}
\begin{equation}
\mathbf{M}_t = \mathbf{G}_i \odot \tanh(\widetilde{\mathbf{M}}_t) + \mathbf{G}_f \odot \mathbf{M}_{t-{win}},
\vspace{-0.07in}
\end{equation}
where ${win}$ is the sliding window size of \emph{NRNM} cell which controls the updating frequency of memory state. $\mathbf{G}_i$ and $\mathbf{G}_f$ are input gate and forget gate respectively to balance the memory information flow from current time step $\widetilde{\mathbf{M}}_t$ and previous memory state $\mathbf{M}_{t-{win}}$. They are modeled by measuring the compatibility between current input feature and previous memory state:
\vspace{-0.05in}
\begin{equation}
\vspace{-0.1in}
\begin{split}
&\mathbf{G}_i = \text{sigmoid} ({\mathbf{W}_{im}\cdot [\mathbf{x}_{t-k+1}, \dots, \mathbf{x}_t, \mathbf{M}_{t-{win}}]}+\mathbf{B}_{im}),\\
& \mathbf{G}_f = \text{sigmoid} ({\mathbf{W}_{fm}\cdot [\mathbf{x}_{t-k+1}, \dots, \mathbf{x}_t, \mathbf{M}_{t-{win}}]}+\mathbf{B}_{fm}), \\
\end{split}
\vspace{-0.05in}
\end{equation}
wherein, $\mathbf{W}_{im}$ and $\mathbf{W}_{fm}$ are transformation matrices while $\mathbf{B}_{im}$ and $\mathbf{B}_{fm}$ are bias terms.

\smallskip\noindent\textbf{Modeling long-range dependencies.}
We aim to capture underlying long-range dependencies in a sequence by a two-pronged strategy: 
\begin{itemize}
	\vspace{-0.06in}
\item We perform non-local operations within each temporal block by \emph{NRNM} cell to capture the full-order interactions locally between different time steps and distill the high-quality memory state. Hence, the local but high-resolution temporal information can be captured.
\vspace{-0.08in}
\item We update the memory state in a gated recurrent manner smoothly when sliding the window of memory block along the temporal domain. It is designed to capture the global temporal dependencies between memory blocks in low resolution considering the potential information redundancy and computational efficiency.
\end{itemize}


\vspace{-0.1in}
\subsection{Sequence Modeling}
\begin{figure}[tb]
	\begin{center}
		\includegraphics[width=0.6\linewidth]{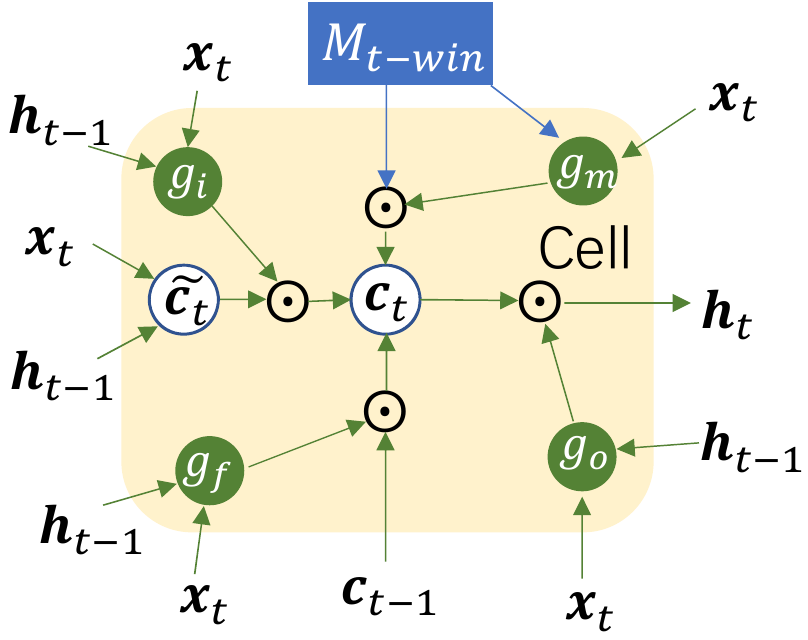}
	\end{center}
	\vspace{-0.1in}
	\caption{The LSTM cell is updated by incorporating the memory state.}
	\label{fig:cell}
	\vspace{-0.15in}
\end{figure}

\vspace{-0.05in}
\label{sec:lstm}
Our \emph{NRNM} can be seamlessly integrated into the LSTM backbone to enhance the power of sequence modeling. Specifically, we incorporate the obtained memory state into the recurrent update of LSTM cell states to help refine its quality as shown in Figure~\ref{fig:cell}:
\vspace{-0.1in}
\begin{equation}
\vspace{-0.1in}
\begin{split}
& \mathbf{v}_m = \text{flatten}(\mathbf{M}_{t-win}) \\
& \mathbf{C}_{t}=\mathbf{g}_{f}\odot \mathbf{C}_{t-1}+\mathbf{g}_{i}\odot\widetilde{\mathbf{C}}_{t}+\mathbf{g}_{m}\odot \mathbf{v}_m,
\label{eqn:lstm_cell}
\end{split}
\end{equation}

where $\mathbf{C}_{t-1}$, $\mathbf{C}_{t}$ and $\widetilde{\mathbf{C}}_{t}$ are previous LSTM cell state, current cell state and candidate cell state  respectively. $\mathbf{v}_m$ is the vector flattened from the memory state $\mathbf{M}_{t-win}$. $\mathbf{g}_f$ and $\mathbf{g}_i$ are the routine forget gate and input gate of LSTM cell to balance the information flow between the current time step and previous step. All $\widetilde{\mathbf{C}}_{t}$, $\mathbf{g}_f$ and $\mathbf{g}_i$ are modeled in a similar nonlinear way as a function of input feature $\mathbf{x}_t$ and previous hidden state $\mathbf{h}_{t-1}$. For instance, the input gate $\mathbf{g}_i$ is modeled as:
\vspace{-0.1in}
\begin{equation}
\vspace{-0.05in}
\mathbf{g}_i = \text{sigmoid}(\mathbf{W}_i \cdot \mathbf{x}_t + \mathbf{U}_i \cdot \mathbf{h}_{t-1}+ \mathbf{b}_i).
\label{eqn:gate}
\end{equation}
In Equation~\ref{eqn:lstm_cell}, we enrich the modeling of the LSTM cell state $\mathbf{C}_{t}$ by incorporating our \emph{NRNM} cell state $\mathbf{M}_t$ via a memory gate $\mathbf{g}_{m}$. The memory gate is constructed as a matrix to control the information flow from the memory state $\mathbf{M}_t$, which is derived by measuring the relevance (compatibility) between current input feature and the memory state:
\vspace{-0.05in}
\begin{equation}
\vspace{-0.05in}
\begin{split}
& \mathbf{g}_{m}=\text{sigmoid} (\mathbf{W}_{m}\cdot \mathbf{x}_{t}+\mathbf{U}_{m}\cdot \text{flatten}(\mathbf{M}_{t-win})+\mathbf{b}_{m}),
\end{split}
\end{equation}
where $\mathbf{W}_{m}$ and $\mathbf{U}_{m}$ are transformation matrices and $\mathbf{b}_m$ is the bias term.

The newly constructed cell state $\mathbf{C}_{t}$ are further used to derive the hidden state $\mathbf{h}_t$ of the whole sequence model prepared for the final prediction:
\vspace{-0.05in}
\begin{equation}
\vspace{-0.05in}
\mathbf{h}_{t}=\mathbf{g}_{o}\odot \tanh(\mathbf{C}_{t}),
\label{eqn:hidden}
\end{equation}
where $\mathbf{g}_o$ is the output gate which is modeled in a similar way to the input gate in Equation~\ref{eqn:gate}.

\subsection{End-to-end Parameter Learning}
\vspace{-0.05in}
The memory state of our \emph{NRNM} for current block is learned 
based on the hidden states within this block of the LSTM backbone while the obtained memory state is in turn leveraged to refine 
the hidden states in future time steps. Hence, our \emph{NRNM} and the LSTM backbone are integrated seamlessly and refine each other alternately.

The learned hidden representations $\{\mathbf{h}_t\}_{t=1, \dots, T}$ in Equation~\ref{eqn:hidden} for a sequence with length $T$ can be used for any sequence prediction task such as step-wise prediction (like language modeling) or sequence classification (like action classification). In subsequent experiments, we validate our model in two tasks of sequence classification with different modalities: action recognition and sentiment analysis. Below we present the loss function for training our model for sequence classification, but it is straightforward to substitute the loss function to adapt our model to tasks of step-wise prediction.

Given a training set $\mathcal{D}=\{\mathbf{x}_{1, \dots, T^n}^{n}, y^{n}\}_{n=1, \dots, N}$ containing $N$ sequences of length $T^n$ and their associated labels $y^{n}$. We learn our \emph{NRNM} and the LSTM backbone jointly in an end-to-end manner by minimizing the conditional negative log-likelihood of the training data with respect to the parameters:
\vspace{-0.1in}
\begin{equation}
\vspace{-0.05in}
\mathcal{L} = -\sum_{n=1}^N \log P(y^n | \mathbf{x}_{1, \dots, T^n}^n),
\end{equation}
where the probability of the predicted label $y^n$ among $K$ classes is calculated by the hidden state in the last time step:
\vspace{-0.05in}
\begin{equation}
\vspace{-0.05in}
P(y^n | \mathbf{x}_{1, \dots, T^n}^n) = \frac{\text{exp}(\mathbf{W}^\top_{y^n} \mathbf{h}_{T^n}+\mathbf{b})}{\sum_{i=1}^K\text{exp}(\mathbf{W}_i^\top \mathbf{h}_{T^n}+\mathbf{b}_i)}.
\end{equation}
Herein, $\mathbf{W}^\top$ and $\mathbf{b}$ is the parameters for linear transformation  and bias term.

\vspace{-0.1in}
\section{Experiments on Action Recognition}
\vspace{-0.05in}
To evaluate the performance of our proposed \emph{NRNM} model, we first consider the task of action recognition in which the temporal dependencies between frames in a video are the most discriminative cues.

\begin{figure*}[tbp]
	\centering                                              
	\subfigure[]{
		\label{fig:ablation:block}                  
		\begin{minipage}[t]{0.25\linewidth}
			\centering                                                        
			\includegraphics[width=1.0\textwidth]{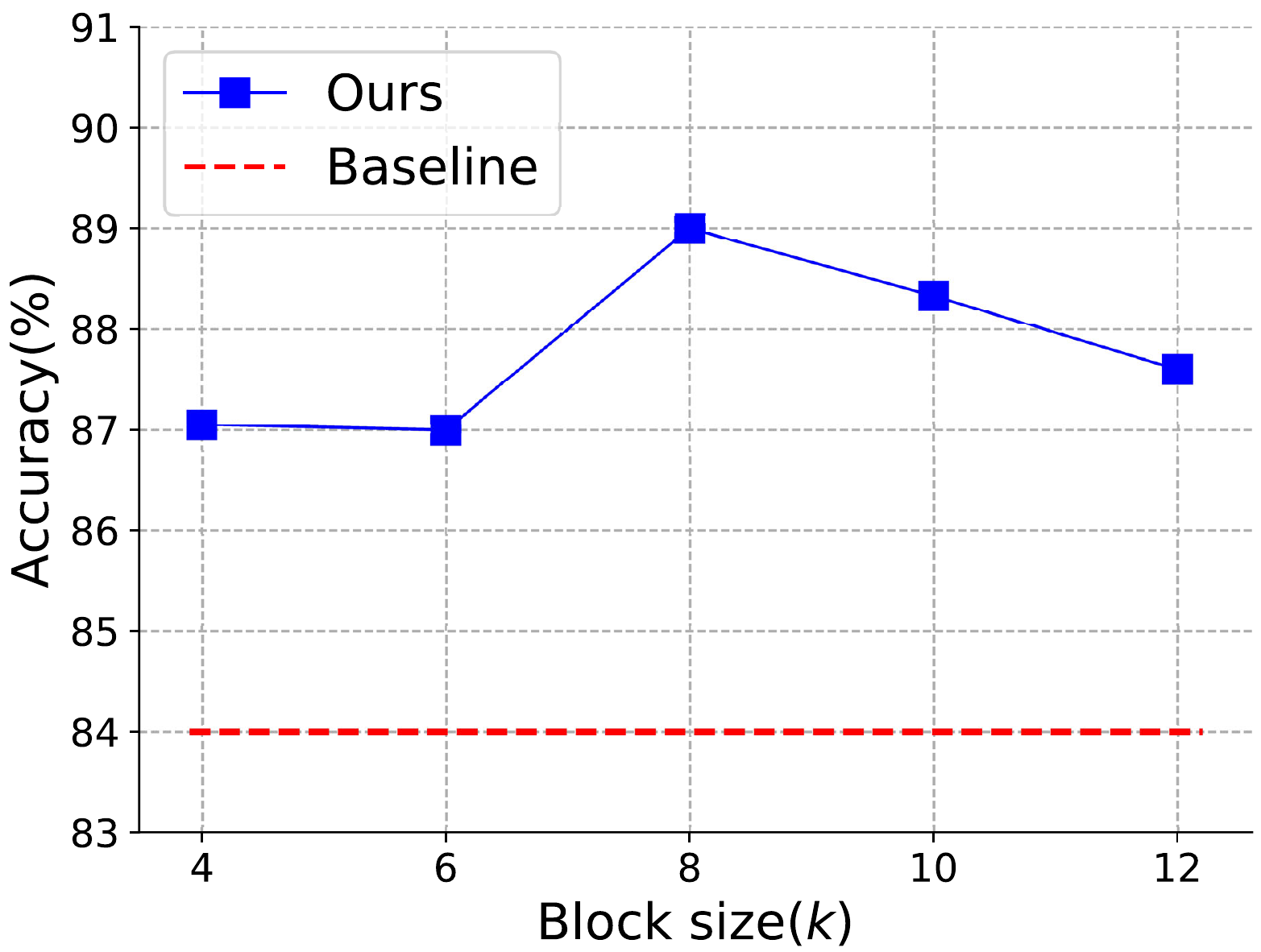}              
	\end{minipage}}
	\subfigure[]{
		\label{fig:ablation:memory}                    
		\begin{minipage}[t]{0.25\linewidth}
			\centering                                                      
			\includegraphics[width=1.0\textwidth]{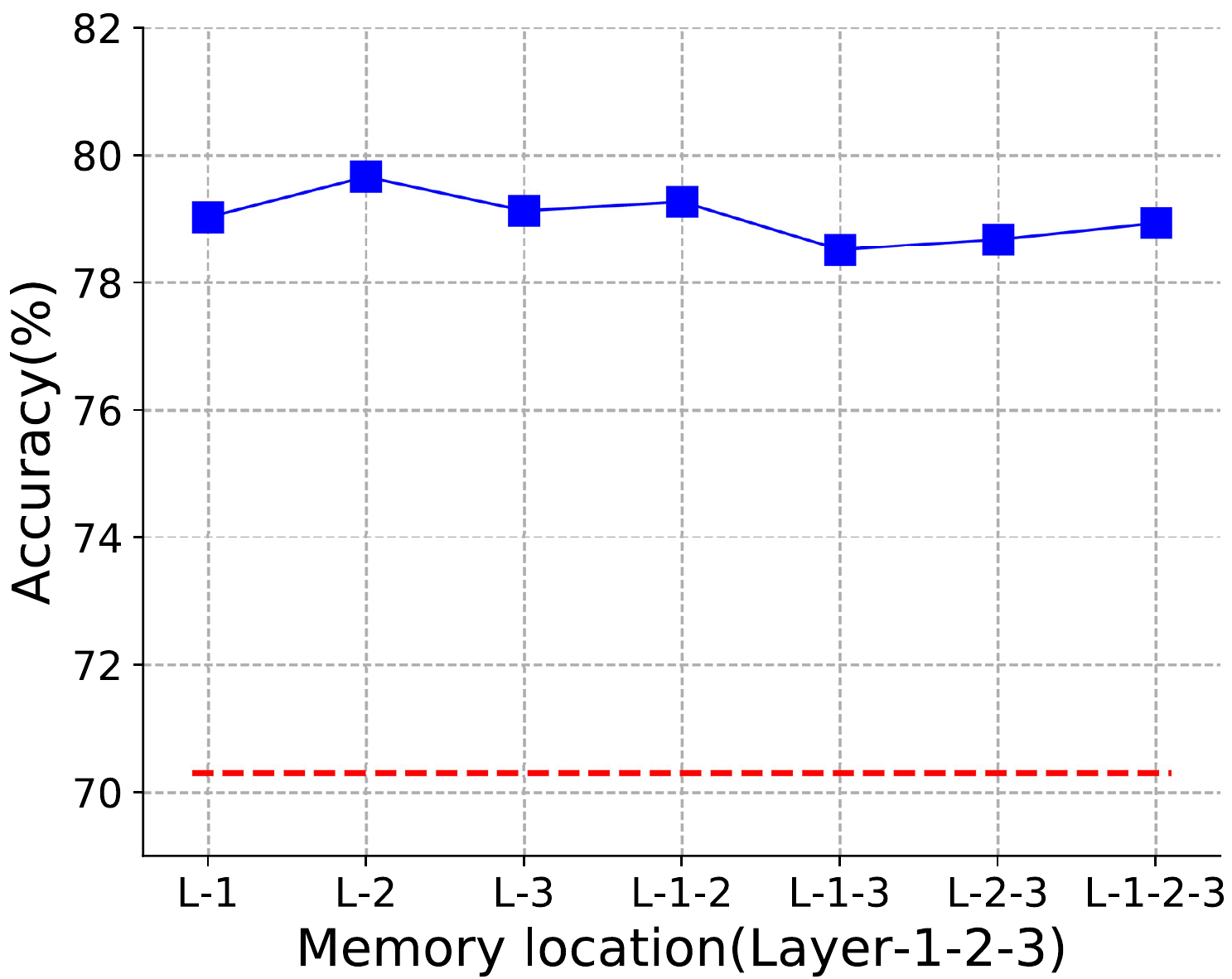}              
	\end{minipage}}
	\subfigure[]{  
		\label{fig:ablation:sliding}             
		\begin{minipage}[t]{0.25\linewidth}
			\centering                                                        
			\includegraphics[width=1.0\textwidth]{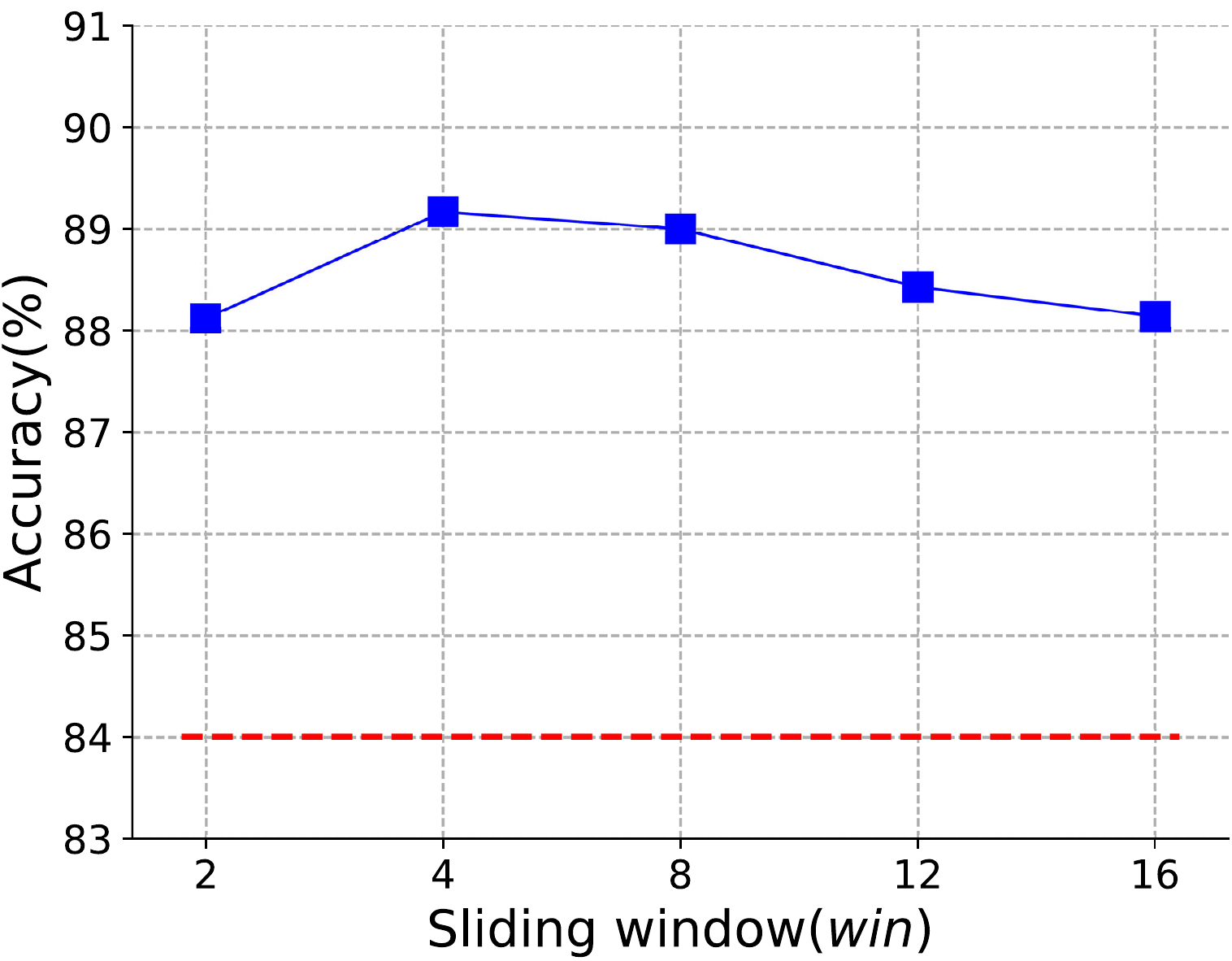}    
	\end{minipage}}
	\caption{Ablation study of \emph{NRNM} on NTU dataset by exploring the effect of (a) the block size $k$ , 
	(b) the integrated location of \emph{NRNM} on the LSTM backbone and (c) the sliding window size $win$. The performance of the baseline (a standard LSTM) is presented for reference.}                  
	\vspace{-0.15in}   
	\label{fig:ablation}               
\end{figure*}
\vspace{-0.05in}
\subsection{Dataset and Evaluation Protocol}
\vspace{-0.05in}
We evaluate our method on the NTU dataset\cite{shahroudy2016ntu} which is currently the largest action recoginition dataset. It is a RGB+D-based dataset containing 56,880 video sequences and 4 million frames collected from 40 distinct subjects. The dataset includes 60 action categories. 3D skeleton data (i.e. 3D coordinates of 25 body joints) is provided using Microsoft Kinect. 

In our experiments, we opt for NTU dataset using only 3D skeleton joint information rather than Kinetics~\cite{kay2017kinetics} based on RGB information for action recognition since single-frame RGB information already provides much implication for action recognition and weakens the importance of temporal dependencies~\cite{qiu2017learning}.
Dropping RGB-D information enforces our model to recognize actions relying on temporal information of joints.

Two standard evaluation metrics are provided in~\cite{shahroudy2016ntu}: Cross-Subject (CS) and Cross-View (CV). CS evaluation splits 40 subjects equally into training and test sets consisting of 40,320 and 16,560 samples respectively. In CV evaluation, samples of camera 1 are used for testing and samples from cameras 2 and 3 for training. We report both metrics for performance evaluation. 

\vspace{-0.05in}
\subsection{Implementation}
\vspace{-0.05in}
Our \emph{NRNM} is built on a 3-layer LSTM backbone.  The number of hidden units of all recurrent networks mentioned in this work 
(vanila-RNN, GRU, LSTM) is tuned on a validation set by selecting the best configuration from the option set $\{128, 256, 512\}$. 
We employ 4-head attention scheme in practice.
The size of memory state is set to be same as the combined size of input hidden states, i.e., the dimensions are $[\text{block size } (k) /\text{stride } (s),  \text{dim} (\mathbf{h}_t)]$. Following Tu et al.~\cite{tu2018spatial}, Zoneout~\cite{krueger2016zoneout} is employed for network regularization. The dropout value is set to 0.5 to prevent potential overfitting. Adam~\cite{kingma2014adam} is used with the initial learning rate of 0.001 for gradient descent optimization. 
\begin{figure}[htb]
	\begin{center}
		\includegraphics[width=0.9\linewidth]{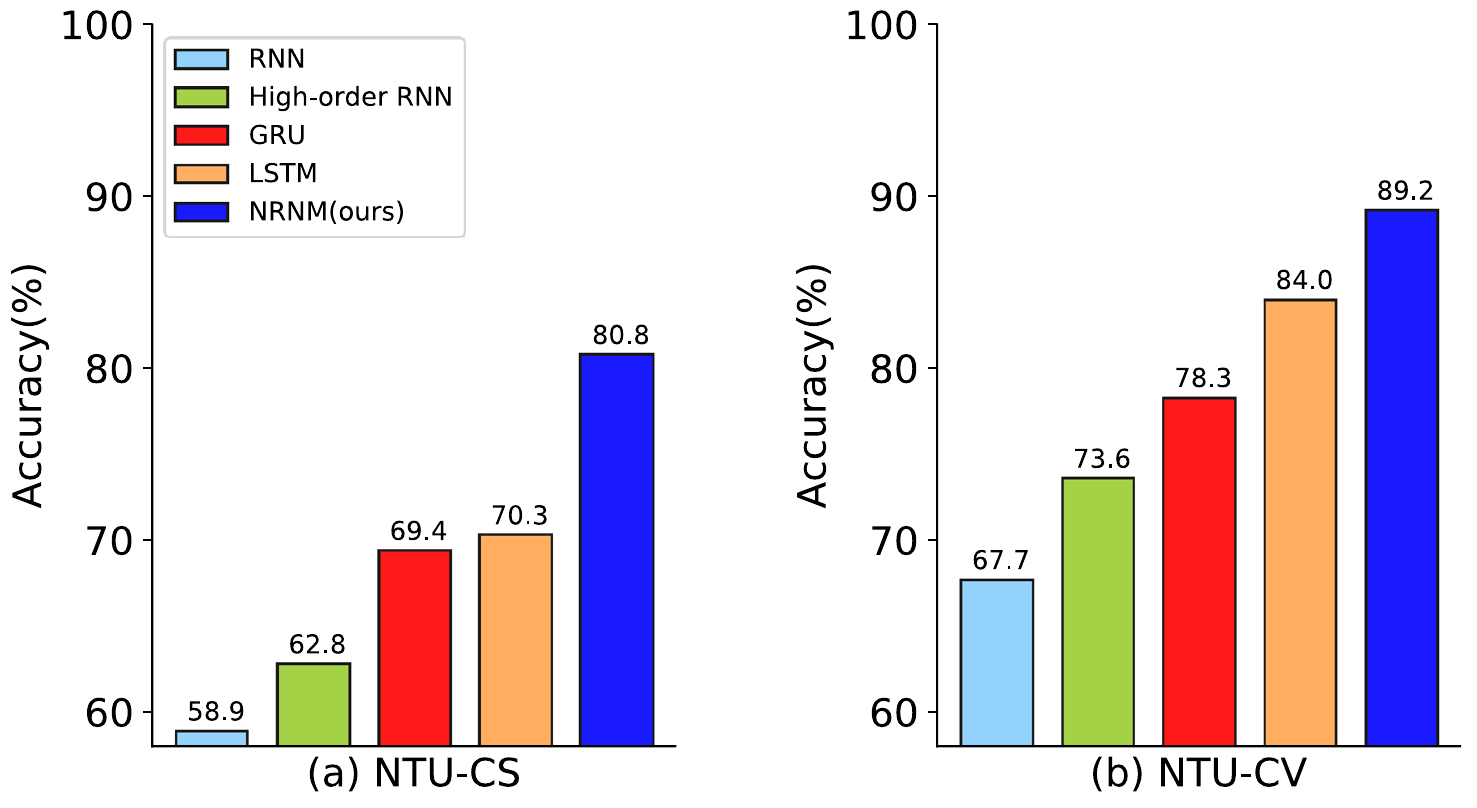}
	\end{center}
\vspace{-0.1in}
	\caption{Comparison of our model with other basic recurrent models in terms of classification accuracy ($\%$) on NTU dataset in both Cross-Subject (CS) and Cross-View (CV) metrics.}
	\vspace{-0.2in}
	\label{fig:ntu_base}
\end{figure}

\begin{figure*}[!tb]
	\begin{center}
		\includegraphics[width=0.8\linewidth]{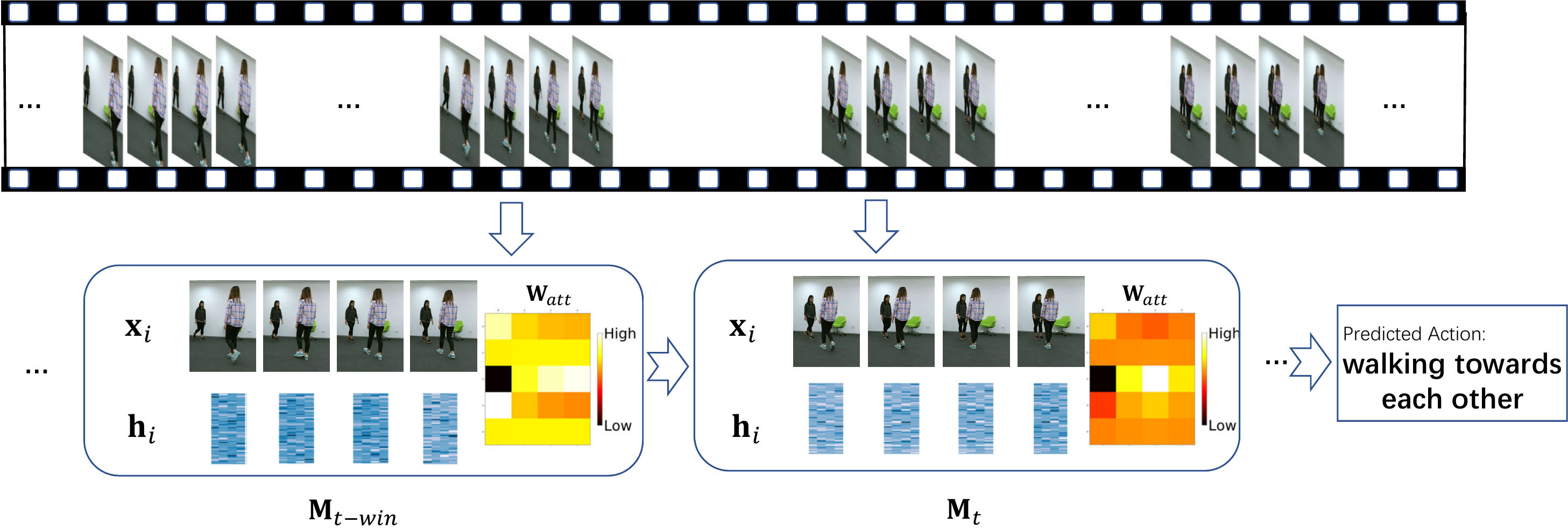}
	\end{center}
	\vspace{-0.15in}
	\caption{Visualization of an example with labeled action ``walking towards each other".  Our model is able to correctly recognize it while LSTM misclassifies it as ``punching/slapping other person". The temporal variations of relative distance between two persons are key to recognize the action. Our model can successfully capture it while LSTM fails. Two blocks of memory states and the attention weights $\mathbf{W}_{att}$ in Equation~\ref{eqn:transform} are visualized.  }
	\label{fig:vis_action}
	\vspace{-0.2in}
\end{figure*}

\vspace{-0.05in}
\subsection{Investigation on NRNM}
\vspace{-0.05in}
We first perform experiments to investigate our proposed \emph{NRNM} systematically.

\smallskip\noindent\textbf{Effect of the block size $k$.}
We first conduct experiments on NTU dataset to investigate the performance of \emph{NRNM} as a function of the block size.
Concretely, we evaluate our method using an increasing number of block sizes: 4, 6, 8, 10, and 12 while fixing the other hyper-parameters.

Figure~\ref{fig:ablation:block} shows that the accuracy initially increases as the increase of the block size, 
which is reasonable since larger block size allows \emph{NRNM} to incorporate information of more time steps in memory and thus enables \emph{NRNM} to capture longer temporal dependencies. As the block size increases further after the saturated state at the block size of 8, the performance starts to decrease. We surmise that the non-local operations on a long block of sequence result in overfitting on the training data and information redundancy. 

\smallskip\noindent\textbf{Effect of the integrated location of NRNM on the LSTM backbone.}
We next study the effect of integrating the \emph{NRNM} into different layers of the 
3-layer LSTM backbone. Figure~\ref{fig:ablation:memory} presents the results, 
from which we can conclude: 1) Integrating \emph{NRNM} at any layer of LSTM outperforms the standard LSTM; 
2) Only integrating \emph{NRNM} once at one layer performs better than applying \emph{NRNM} at multiple 
layers which would lead to information redundancy and overfitting; 3) Integrating \emph{NRNM} at the middle 
layer achieves the best performance, which is probably because the layer-2 hidden states of LSTM are more 
suitable for \emph{NRNM} to distill information than the low-level and high-level features learned by 
layer-1 and layer-3 hidden states.

 


\smallskip\noindent\textbf{Effect of sliding window size $win$.}
Then we investigate the effect of sliding window size, which is used to control the updating frequency of memory state. Theoretically, too small sliding window size implies much overlap between 
two adjacent memory blocks and thus tends to lead to information redundancy. On the other hand, too large sliding window size results in large non-accessed temporal interval between two adjacent memory blocks and would potentially miss information in the interval.

In this set of experiments, we set block size to 8 time steps, and consider different sliding window size. Figure~\ref{fig:ablation:sliding} reports that the model performs well when the sliding window is around 4 to 8 while the performance decreases at other values, which validates our analysis. 

\smallskip\noindent\textbf{Comparison with LSTM baselines.}
To investigate the effectiveness of our \emph{NRNM}, we compare our model to the basic recurrent models 
including vanila-RNN, GRU, LSTM and high-order RNN on NTU dataset in two evaluation metrics: Cross-Subject (CS) and Cross-View (CV). Figure~\ref{fig:ntu_base} shows
1) all RNNs with memory or gated structure outperforms vanila-RNN and high-order RNN by a large margin, 
which indicates the advantages of memory and gated structure for controlling information flow; 2) high-order RNN 
performs better than vanila-RNN which implies the necessary of the non-local operations since high-order 
connections can be considered as a simple non-local operation in a local area. It is also consistent with the existing conclusions ~\cite{soltani2016higher, zhang2018high}; 3) our \emph{NRNM} outperforms LSTM significantly which demonstrates the superiority of our model over standard LSTM. 

\vspace{-0.05in}
\subsection{Comparison with State-of-the-arts}
\vspace{-0.05in}
In this set of experiments, we compare our model with the state-of-the-art methods for action recognition on 
NTU dataset in both Cross-Subject (CS) and Cross-View (CV) metrics. It should be noted that we do not compare 
with methods which employ extra information or prior knowledge such as joint connections for each part of 
body or human body structure modeling~\cite{si2018skeleton, yan2018spatial}.

Table~\ref{tabel:NTUresult} reports the experimental results. Our model achieves the best performance in both CS and CV metrics, 
which demonstrates the superiority of our model over other recurrent networks, especially those with memory or gated structures. 
While our model outperforms the standard LSTM model substantially, the methods based on LSTM~\cite{song2018spatio, zhang2018adding} 
boost the performance over LSTM by introducing extra attention mechanisms. 

 \begin{table}[!htbp]
	\centering
	\renewcommand{\arraystretch}{1.2}
	\resizebox{0.6\linewidth}{!}{
		\begin{tabular}{c|cc}   
			\toprule
			&CS & CV  \\  
			\midrule       
			HBRNN-L\cite{du2015hierarchical}  & 59.1 & 64.0 \\
			Part-aware LSTM\cite{shahroudy2016ntu} & 62.9 & 70.3 \\
			Trust Gate ST-LSTM\cite{liu2016spatio} & 69.2 & 77.7 \\
			Two-stream RNN\cite{wang2017modeling} & 71.3 & 79.5 \\
			Ensemble TS-LSTM\cite{lee2017ensemble} & 74.6 & 81.3 \\
			VA-LSTM\cite{zhang2017view} & 79.4 & 87.6 \\
			STA-LSTM\cite{song2018spatio}   & 73.4 & 81.2 \\
			EleAtt-LSTM\cite{zhang2018adding} & 78.4 & 85.0 \\
			EleAtt-GRU\cite{zhang2018adding} & 79.8 & 87.1 \\
			\cline{1-3}
			LSTM (baseline)   & 70.3 & 84.0 \\
			\emph{NRNM} (ours)        & \textbf{80.8} & \textbf{89.2} \\
			\bottomrule
		\end{tabular}
	}
	\vspace{0.05in}
	\caption{Classification accuracy (\%) on NTU by different methods in both Cross-Subject (CS) and Cross-View (CV) metrics. 
	}
	\label{tabel:NTUresult}
\end{table}

\smallskip\noindent\textbf{Analysis on model complexity.}
To compare the model complexity between our model and other recurrent baselines and investigate whether the performance gain of our model is boosted by the augmented model complexity,  we evaluate the performance of the recurrent baselines with different model complexities (configurations) in Table~\ref{table:parameters}. Our model substantially outperforms other baselines under optimized configurations, which demonstrates that the performance superiority of our model is not resulted from the increased capacity by the extra parameters. 

\begin{table}[!htbp]
	\vspace{-5pt}
\centering
\renewcommand{\arraystretch}{1.2}
\resizebox{0.65\linewidth}{!}{
  \begin{tabular}{c|c|c}   
	 \toprule
	   &CV(\%) & \#Parameters  \\  
	 \midrule       
	 3-LSTM (256)   & 83.9 &  1.5M\\
	 3-LSTM (512)   & 84.0 &  5.6M\\
	 5-LSTM (512)   & 83.1 &  9.8M\\
	 \midrule
	 3-EleAtt-LSTM (256) & 85.5 &  1.8M\\
	 6-EleAtt-LSTM (256) & 82.7 &  3.8M\\
	 4-EleAtt-LSTM (512) & 83.4 &  8.9M\\
	 \midrule
	 3-EleAtt-GRU (100) & 87.1 &  0.3M\\
	 3-EleAtt-GRU (256) & 85.4 &  1.4M\\
	 5-EleAtt-GRU (256) & 85.0 &  2.5M\\
\midrule
	 \emph{NRNM} (ours)        & \textbf{89.2} &  3.6M\\
	 \bottomrule
  \end{tabular}
}
\vspace{2pt}
\caption{Classification accuracy (\%) on NTU by different methods with different model complexities in Cross-View (CV) metrics. 
Here 3-LSTM (256) denotes the LSTM equipped with 3 hidden layers comprising 256 hidden units. Note that all results are reported from our implementations.}
\label{table:parameters}
  \vspace{-15pt}
\end{table}
\subsection{Qualitative Analysis}
\vspace{-0.05in}
To qualitatively illustrate the advantages of the proposed \emph{NRNM}, 
figure~\ref{fig:vis_action} presents a concrete video example with the action label ``walking towards each other" (groundtruth). 
In this example, it is quite challenging to recognize the action since it can only be inferred by the temporal variations of the 
relative distance between two persons in the scene. Hence, capturing the long-range dependencies is crucial to recognize it. 
The standard LSTM misclassifies it as ``punching/slapping other person'' while our model is able to correctly classify it due to 
the capability to model long-range temporal information by our designed \emph{NRNM}.

Figure~\ref{fig:vis_action} visualizes two blocks of memory states, each of which is learned by 
\emph{NRNM} cell via incorporating information of multiple frames including input features $\mathbf{\text{x}}_i$ and the 
hidden states $\mathbf{\text{h}}_i$ of LSTM backbone. 
To obtain more insights into the non-local operations of \emph{NRNM}, we visualize the attention weights
 $\mathbf{W}_{att}$ in Equation~\ref{eqn:transform} to show that each unit of memory state is calculated by attending to 
all units of source information ($\mathbf{x}_i$ and $\mathbf{\text{h}}_i$).

\vspace{-0.05in}
\section{Experiments on Sentiment Analysis}
\vspace{-0.1in}
\begin{figure*}[!tb]
	\begin{center}
		\includegraphics[width=0.85\linewidth]{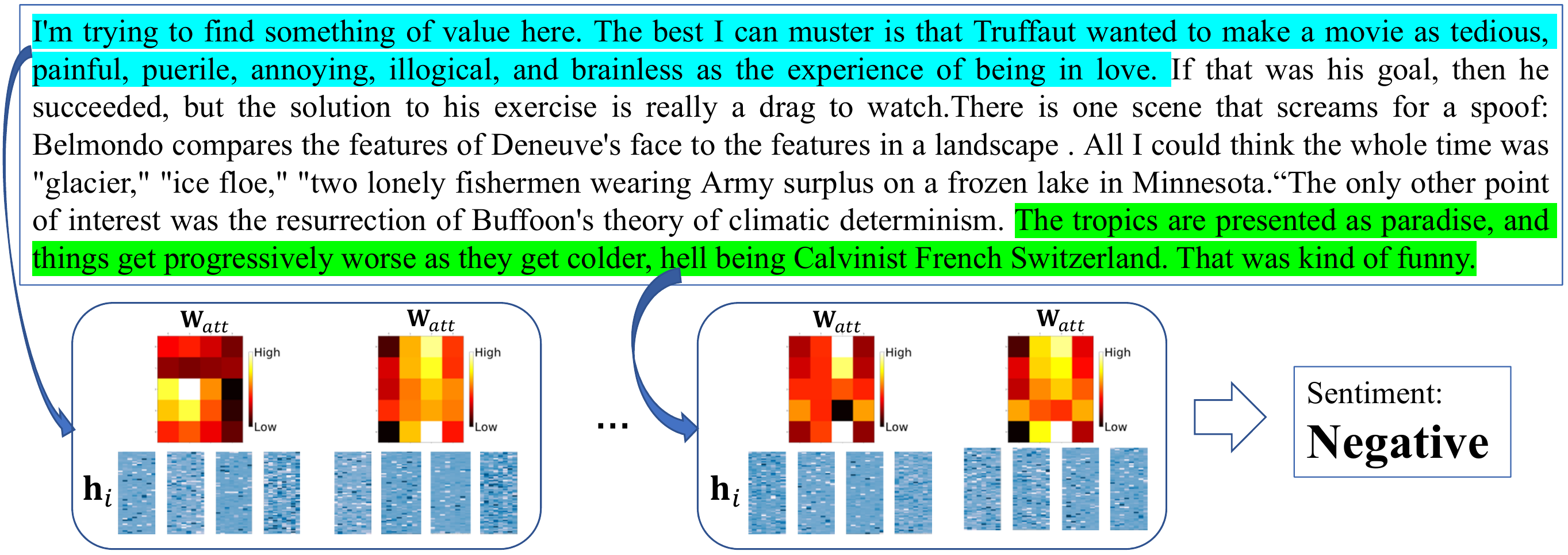}
	\end{center}
	\vspace{-0.15in}
	\caption{Visualization of an example of movie review with the groundtruth label ``negative". Our model is able to correctly classify it while LSTM fails. The last sentence (in green color) which seems positive tends to misguide models. The first sentence is an important cue for negative sentiment, which is hardly captured by LSTM since it is easily forgotten by the hidden state $\mathbf{h}_T$ in the last time step.}
	\label{fig:vis_imdb}
	\vspace{-0.2in}
\end{figure*}

Next we perform experiments on task of sentiment analysis to evaluate our model on the text modality. Specifically, we aim to identify online movie reviews as positive or negative, which is a sequence classification problem.  

\vspace{-0.1in}
\subsection{Dataset and Evaluation Protocol}
\vspace{-0.05in}
We use the IMDB Review dataset   \cite{maas2011learning} which is a stardard benchmark for sentiment analysis. It contains 50,000 labeled reviews among which 
25,000 samples are used for training and the rest for testing. The average length of reviews is 241 words and the maximum length is 
2526 words\cite{dai2015semi}. Note that the IMDB dataset also provides additional 50,000 unlabeled reviews, which are used by 
several customized semi-supervised learning methods \cite{dai2015semi,dieng2016topicrnn,johnson2016supervised,miyato2016adversarial,radford2017learning}. Since we only use labeled data for supervised training, we compare our methods with those methods based on supervised learning using the same set of training data for a fair comparison.

The torchtext~\footnote{\url{https://github.com/pytorch/text}} is used for data preprocessing. Following the training strategy in Dai et al.~\cite{dai2015semi}, we pretrain a language model for extracting word embeddings. 

\vspace{-0.05in}
\subsection{Comparison with LSTM Baselines}
\begin{figure}[htb]
	\begin{center}
		\includegraphics[width=0.4\linewidth]{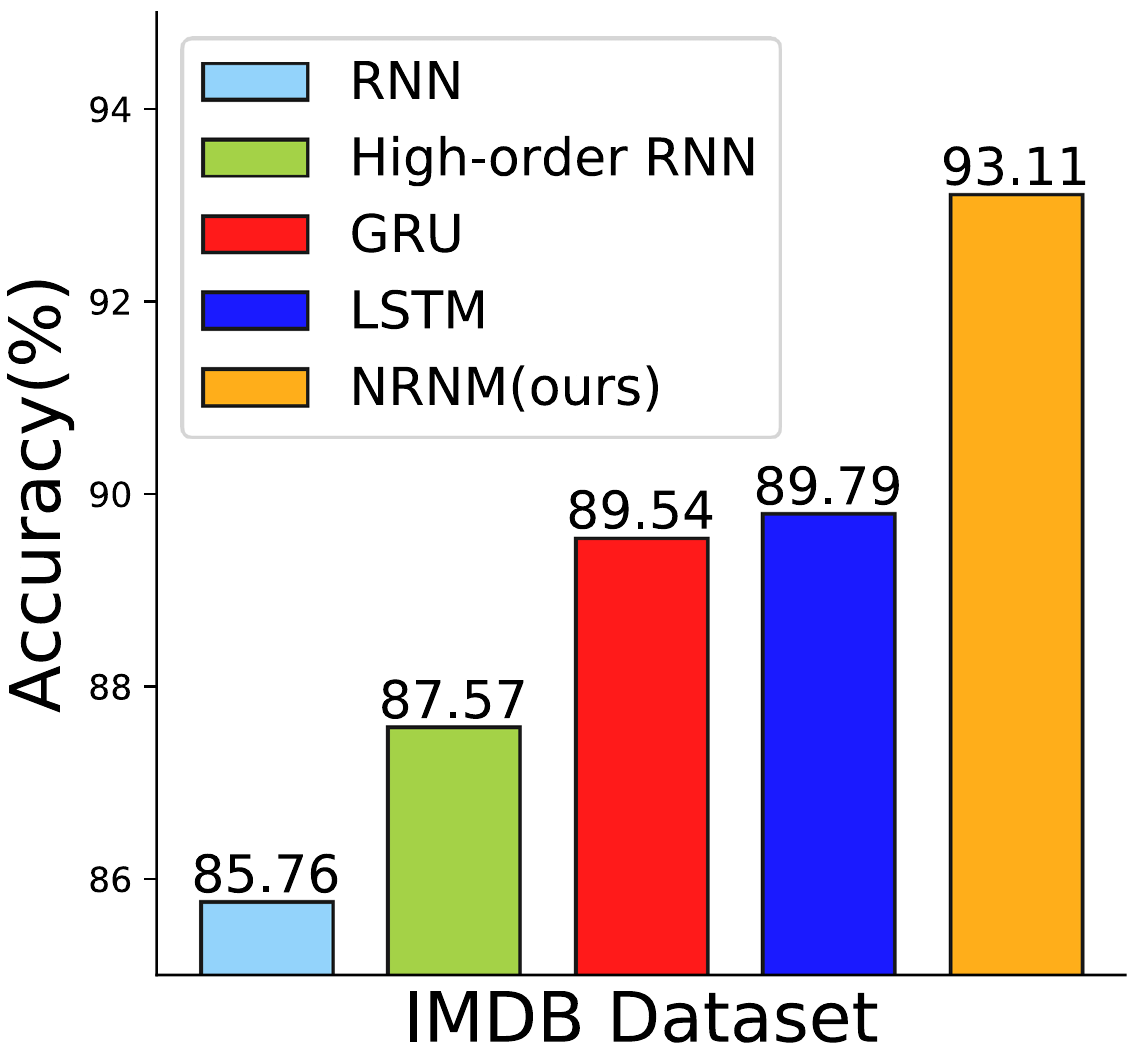}
	\end{center}
\vspace{-0.15in}
	\caption{Comparison of our model with other basic recurrent models in terms of classification accuracy ($\%$) on IMDB dataset.}
	\vspace{-0.15in}
	\label{fig:imdb}
\end{figure}

\vspace{-0.05in}
We first conduct a set of experiments to compare our model to the basic recurrent networks 
including vanila-RNN, GRU, LSTM and high-order RNN. 
Figure~\ref{fig:imdb} shows that our model outperforms all other baselines significantly which reveals the remarkable 
advantages of our \emph{NRNM}. Besides, while LSTM and GRU perform much better than vanila-RNN, high-order  
RNN also boosts the performance by a large margin compared to vanila-RNN. 
It again demonstrates the benefits of high-order connections which are a simple form of non-local operations in local area.
\vspace{-0.1in}

\vspace{-0.10in}
\subsection{Comparison with the State-of-the-arts}
\vspace{-0.05in}
Next we compare our \emph{NRNM} with the state-of-the-art methods including LSTM\cite{xia2018model}, oh-CNN~\cite{johnson2014effective} and oh-2LSTMp\cite{johnson2016supervised} which learn the word embeddings by customized CNN or LSTM instead of using existing pretrained word embedding vocabulary, DSL~\cite{xia2018model} and MLDL\cite{xia2018model} which perform a dual learning between language modeling and sentiment analysis,
GLoMo\cite{yang2018glomo} which is a transfer learning framework, 
and BCN+Char+CoVe\cite{mccann2017learned} which trains a machine translation model to encode the word embeddings to improve the performance of sentiment analysis.  
 

Table~\ref{table:IMDB} shows that our model achieves the best performance among all methods. It is worth mentioning that our model even performs better than GLoMo\cite{yang2018glomo} and BCN+Char+CoVe\cite{mccann2017learned}, which employ additional data for either transfer learning or training a individual machine translation model.



\begin{table}[!htbp]
    \centering
	\renewcommand{\arraystretch}{1.2}
\resizebox{0.5\linewidth}{!}{
    \begin{tabular}{l|c}   
    \toprule
    Methods & Accuracy  \\  
    \midrule    
    LSTM~\cite{xia2018model}  & 89.9 \\
    MLDL~\cite{xia2018model}  & 92.6 \\
    GLoMo~\cite{yang2018glomo}  & 89.2 \\
    oh-2LSTMp~\cite{johnson2016supervised}  & 91.9 \\
        DSL~\cite{xia2018model} & 90.8\\
    oh-CNN~\cite{johnson2014effective} & 91.6\\
    BCN+Char+CoVe~\cite{mccann2017learned}  & 92.1 \\
\midrule
    LSTM (baseline)   & 89.8  \\
   \emph{NRNM} (ours)           & \textbf{93.1}   \\
    \bottomrule
    \end{tabular}
}
\vspace{0.05in}
    \caption{Classification accuracy (\%) on IMDB dataset by different methods. }
        \label{table:IMDB}
        \vspace{-0.18in}
\end{table}

\vspace{-0.05in}
\subsection{Qualitative Analysis}
\vspace{-0.05in}
Figure~\ref{fig:vis_imdb} illustrates an example of sentiment analysis from IMDB dataset. 
This example of movie review is fairly challenging since the last sentence of the review seems 
to be positive which is prone to misguide models, especially when we use the hidden state of last 
time step $\mathbf{h}_T$ for prediction. Our model is able to correctly classify it as ``negative" 
while LSTM fails.  We also visualize the attention weights of non-local operations ($\mathbf{W}_{att}$ 
Equation~\ref{eqn:transform}) in two blocks of \emph{NRNM} states to show the attendance of each information 
units of source information for calculating the \emph{NRNM} states. The first memory block corresponds to the first sentence which is an important cue of negative sentiment while the second memory block corresponds to the last sentence.

\vspace{-0.1in}
\section{Conclusion}
\vspace{-0.05in}
In this work, we have presented the Non-local Recurrent Neural Memory (\emph{NRNM}) 
for supervised sequence modeling. We perform non-local operations within each memory block to model 
full-order interactions between non-adjacent time steps and model the global interactions between memory blocks in a gated recurrent manner. Thus, the long-range temporal dependencies are captured. 
Our method achieves the state-of-the-art performance for tasks of action recognition and 
sentiment analysis. 


{\small
\bibliographystyle{ieee}
\bibliography{egbib}
}

\end{document}